# Toward Reliable and Explainable Nail Disease Classification: Leveraging Adversarial Training and Grad-CAM Visualization

Farzia Hossain
*Dept. of CSE*
*East West University*
Dhaka, Bangladesh
farzimh12@gmail.com

Samanta Ghosh
*Dept. of CSE*
*East West University*
Dhaka, Bangladesh
Samantaewu28@gmail.com

Shahida Begum
*Dept. of CSE*
*East West University*
Dhaka, Bangladesh
imashahida92@gmail.com

B. M. Shahria Alam
*Dept. of CSE*
*East West University*
Dhaka, Bangladesh
bmshahria@gmail.com

Mohammad Tahmid Noor
*Dept. of CSE*
*East West University*
Dhaka, Bangladesh
tahmidnoor770@gmail.com

Md Parvez Mia
*Dept. of CSE*
*East West University*
Dhaka, Bangladesh
mdparvezmia999@gmail.com

Nishat Tasnim Niloy
*Dept. of CSE*
*East West University*
Dhaka, Bangladesh
nishat.niloy@ewubd.edu

*Abstract*— **Human nail diseases are gradually observed over all age groups, especially among older individuals, often going ignored until they become severe. Early detection and accurate diagnosis of such conditions are important because they sometimes reveal our body's health problems. But it is challenging due to the inferred visual differences between disease types. This paper presents a machine learning-based model for automated classification of nail diseases based on a publicly available dataset, which contains 3,835 images scaling six categories. In 224×224 pixels, all images were resized to ensure consistency. To evaluate performance, four well-known CNN models—InceptionV3, DenseNet201, EfficientNetV2, and ResNet50 were trained and analyzed. Among these, InceptionV3 outperformed the others with an accuracy of 95.57%, while DenseNet201 came next with 94.79%. To make the model stronger and less likely to make mistakes on tricky or noisy images, we used adversarial training. To help understand how the model makes decisions, we used SHAP to highlight important features in the predictions. This system could be a helpful support for doctors, making nail disease diagnosis more accurate and faster.**

## I. INTRODUCTION

Nails are one of the important organs of our body. They protect our fingertips. Changes in nail color, texture, or growth can be the signs of diseases such as fungal infections, psoriasis, or melanoma etc. These diseases can be fatal because they can be spread to other parts of the body. The infections may also enter the bloodstream. So, nail disease detection is an important issue for our body. But there is a small number of doctors who use an automated disease detector in our country. They use traditional technologies to detect nail diseases. From this point of necessity, we started the project to develop an automated nail disease detector which will support dermatologists by providing AI-driven diagnostic recommendations, prioritizing patient cases based on urgency, and providing visual representations such as heatmaps to assist clinical decision-making

The study by Regin et al. shows a method for detecting and classifying nail diseases by using deep learning models [1]. The system processes the nail images concerning colour variation and deformation. They used a nail set with selectively labelled conditions and a Weka that pre-processes characteristics. The CNN model is used to extract image features. The study also shows a comparison of CNN performance with traditional machine learning models.

The research by Bang et al. shows an AI-based system to predict nail diseases. The InceptionV3 model is used in this study [2]. The system emphasized on analyzing changes in the color of the nail. By processing the images, the model tries to detect the disease. The study also shows the importance of early disease detection. Classifying nail diseases is focused on the study by Thakur and Kaur [3]. The research focused on the role of deep learning in developing reliable, automated diagnostic tools for nail disease classification.

In this study, DenseNet201, EfficientNetV2, InceptionV3, and ResNet50 are used for nail disease classification. These models are familiar for their high accuracy, efficient feature extraction, and strong performance in medical image analysis. These features allow the models to be suitable for detecting various nail images associated with different diseases.
Research Questions:

- Can a transfer learning model optimize the performance of nail disease detection with limited data?
- To what extent do deep learning models classify various nail diseases from clinical images with high accuracy?

## II. RELATED WORKS

Kemenes et al. [4] introduced a deep learning system using BEiT to evaluate nail psoriasis based on mNAPSI scores. They trained it on 4,400 nail images and tested it with 929 more, achieving strong performance (AUROC: 0.86 train, 0.80 test). The model's predictions closely matched expert assessments, with correlation scores of 0.94 and 0.92. Nails were isolated using MediaPipe, and involving multiple expert reviewers improved scoring reliability. The tool shows promise for clinical use and patient self-monitoring.

Horikawa et al. [5] proposed a deep learning tool called the "NAPSI calculator" to automatically measure nail psoriasis severity from clinical images. The system first detects nails and then calculates the Nail Psoriasis Severity

Index (NAPSI), reducing errors caused by human observers. It showed 83.9% accuracy, better than both dermatology residents 65.7% and board-certified dermatologists 73.0%. The tool was trained using over 2900 nail images and can work on a basic computer setup. This method helps doctors quickly and reliably assess nail psoriasis in clinics.

Folle et al. [6] developed a deep learning system designed to automatically evaluate nail psoriasis severity using the modified NAPSI (mNAPSI) score. They trained their model on over 1,100 nail images from 177 patients, using a transformer-based architecture called BEiT. The system demonstrated strong results, with an 88% AUROC and a 90% match with expert assessments at the patient level. It uses hand key point detection to extract nail regions and then scores them through a dedicated app. The full platform is accessible online for both clinical and research use.

Jansen et al. [7] created a deep learning tool using a U-NET model to help detect fungal infections (onychomycosis) in nail tissue slides. The system was trained on 664 images and compared against 11 dermatopathologists. It achieved 94% sensitivity and 86.5% accuracy, performing similarly to expert doctors. The model highlights fungal regions, helping pathologists review slides faster. This AI tool can assist in early, accurate fungal nail infection diagnosis in clinical practice.

Indi et al. [8] established a system that analyzes nail images to detect early-stage diseases using color features. The system extracts average RGB values from nail photos and compares them with a trained dataset using the C4.5 algorithm in WEKA. Images are captured with a mobile or digital camera, and predictions are made based on color variations. Tested on five diseases, the system achieved around 65% accuracy and helps reduce unnecessary medical tests.

Hamim et al. [9] built a deep learning system to classify nail diseases using image data and transfer learning. They used CNN architectures (MobileNetV2, VGG16, VGG19) to detect three types of nail conditions: bluish nails, red puffy nails, and yellow fungal nails. MobileNetV2 gave the best accuracy 92% after training on over 3,400 images using advanced preprocessing and augmentation techniques. This system speeds up early diagnosis and helps reduce testing time in the medical field..

## III. METHODOLOGY

This Fig. 1 below illustrates the complete workflow of the suggested classification model. It starts with gathering images from the Kaggle repository. These unprocessed visuals are initially sent through a preparation phase, where they undergo resizing and normalization. The data Augmentation process is applied to increase the number of images and improve the color quality of the images. Moreover, we divide the dataset into training, validation, and testing sets in a 70:20:10 ratio for proper evaluation of the models. Furthermore, these processed images are then divided into batches and fed into four different convolutional neural network (CNN) models: InceptionV3, DenseNet201, EfficientNetV2, and ResNet50. During training, adversarial examples are also generated and used to improve the model's robustness. After training, Explainable AI like Grad-CAM is applied to interpret the model's predictions. The visualization makes the study clearer and understandable.

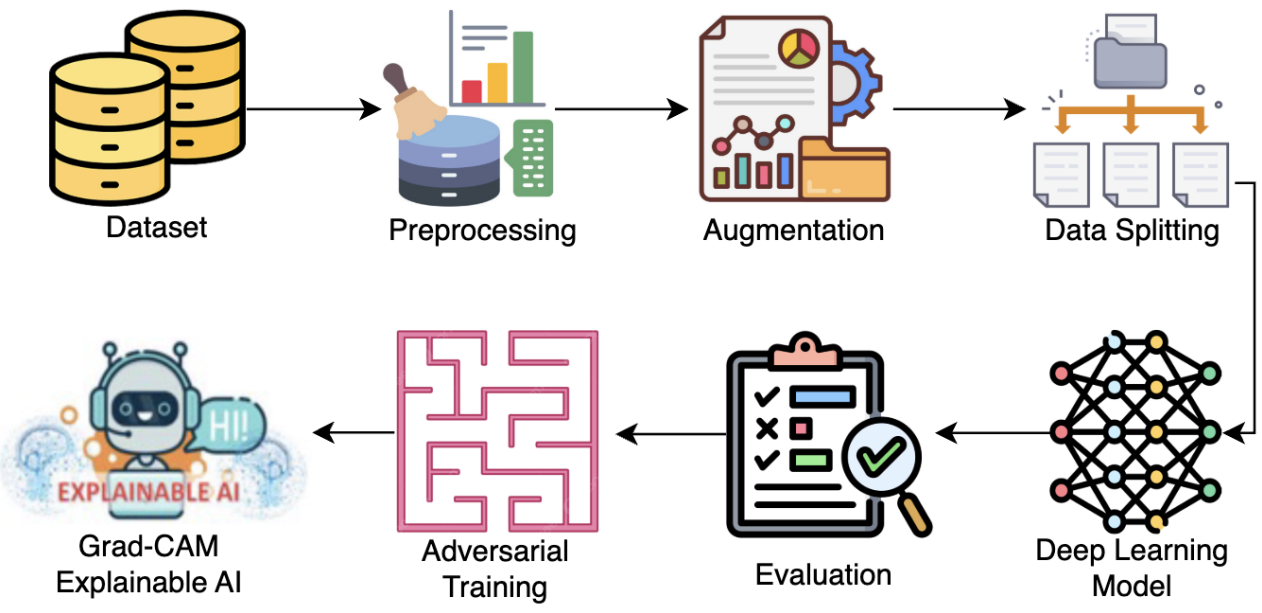

Fig. 1. Methodology

In this research, we used a dataset of human nail diseases shown in Fig. 2, which is publicly available on the Kaggle dataset [10]. The dataset consist of a total of 3,835 nail images in JPG format. The images are classified into six categories: Acral Lentiginous Melanoma, Onychogryphosis, Blue Finger, Clubbing, Pitting, and Healthy Nail. First, Acral Lentiginous Melanoma is a skin cancer that appears on the nails, which is seen as rare. Second, Onychogryphosis nail condition where nails seem thickened, curved, and claw-like. Third, In Blue finger disease, nails appear blue or purple, which is caused by poor oxygen supply to the blood. Fourth, Swollen finger nails that curve around the fingertip are called clubbing. Fifth, Pitting nail disease is sometimes observed by the small dents or pits in the nail surface. Finally, in this dataset, Healthy nails are mainly normal nails with no disease. Each picture was resized to 224×224 pixels to ensure uniform input shape across the different architectures. This ensures efficient transfer learning while maintaining important image features. GPU optimization during training, the data was then organized into batches of size 32.

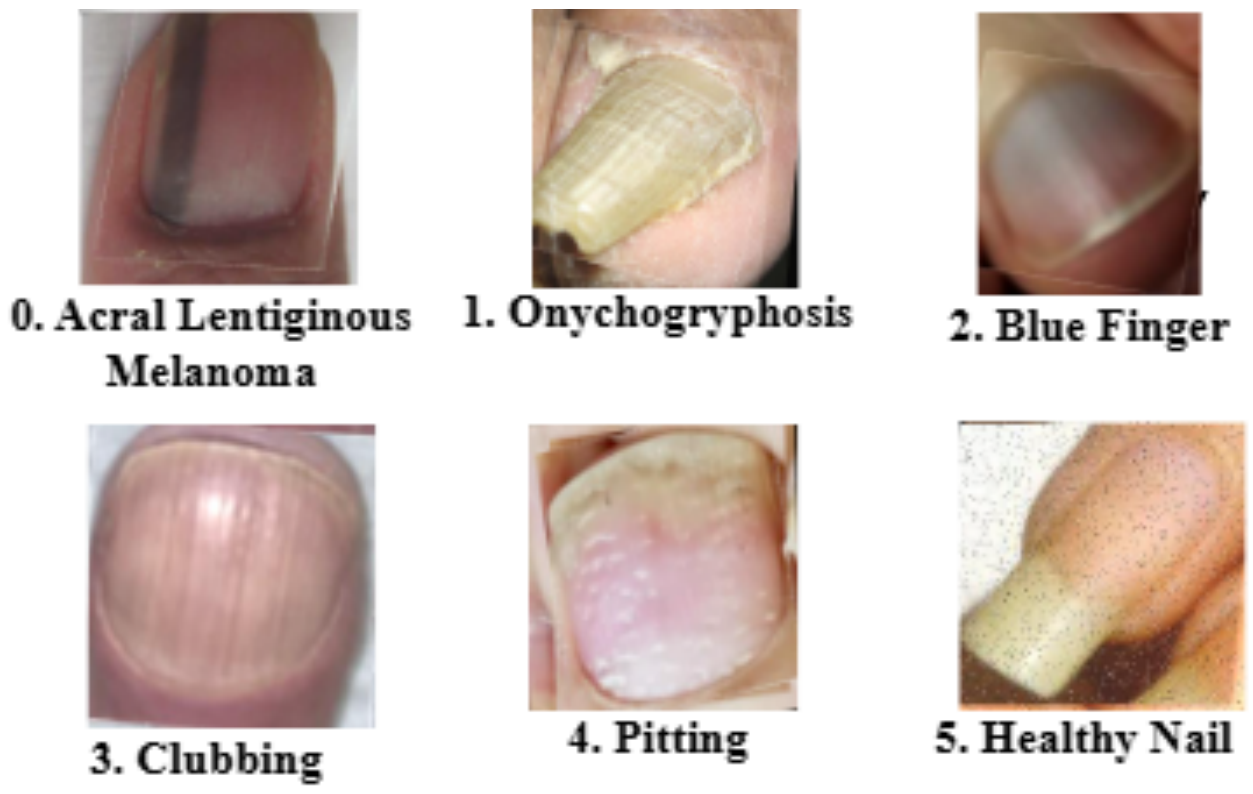

Fig. 2. Dataset Sample

### A. InceptionV3

To reduce computational complexity within the same layer, InceptionV3 is a great choice. It is an Advanced convolutional neural network designed for image classification. It interposes multiple convolutional filters of varying sizes (1×1, 3×3, and 5×5) in parallel within the same layer. Fig. 3 shows the architecture of the InceptionV3 model. To reduce the image size while keeping key features, the model begins with convolution and max pooling layers. Instead of large 5×5 filters, it uses two 3×3 filters in a row, and 3×3 filters are split into 1×3 and 3×1 to lower computation. Each block runs multiple filters in parallel, including 1×1, 3×3, and pooled features, helping the network capture both detailed edges and deeper patterns. A single dense layer is used before the final SoftMax output to classify the image into the 6 nail disease categories.

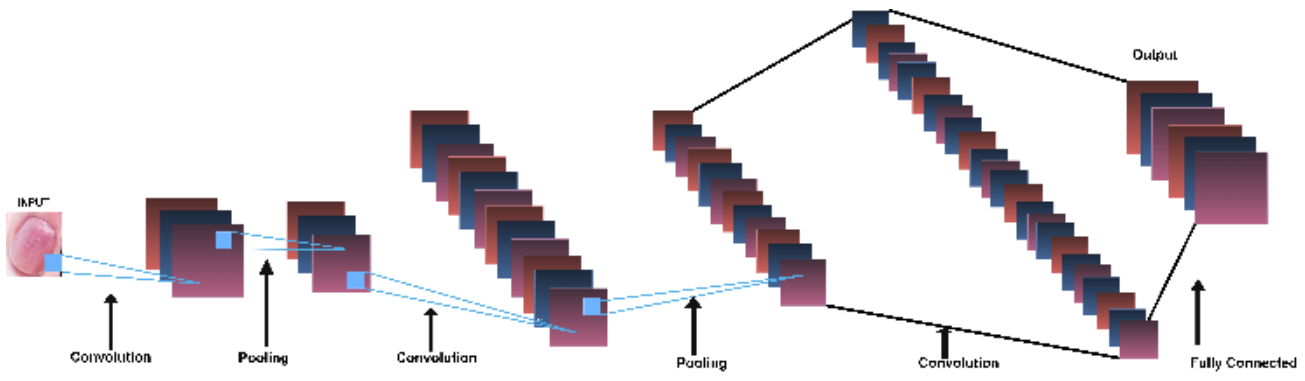

Fig. 3. Proposed InceptionV3 architecture

## B. DenseNet201

DenseNet201 uses a distinctive design where each layer gets input from all earlier layers and shares its output with later ones. It has 201 layers and about 20 million parameters but stays efficient because of how Optimally it's built. Fig. 4 shows, the network begins with a 7×7 convolution and a 3×3 max pooling. "Each dense block has layers that apply batch normalization, ReLU, 1×1 convolution, and 3×3 convolution. Layers produce new feature maps (growth rate of 32), which are stacked. Transition layers help reduce size using pooling and 1×1 convolution. These layers minimize the number of feature maps and spatial resolution to control model size.

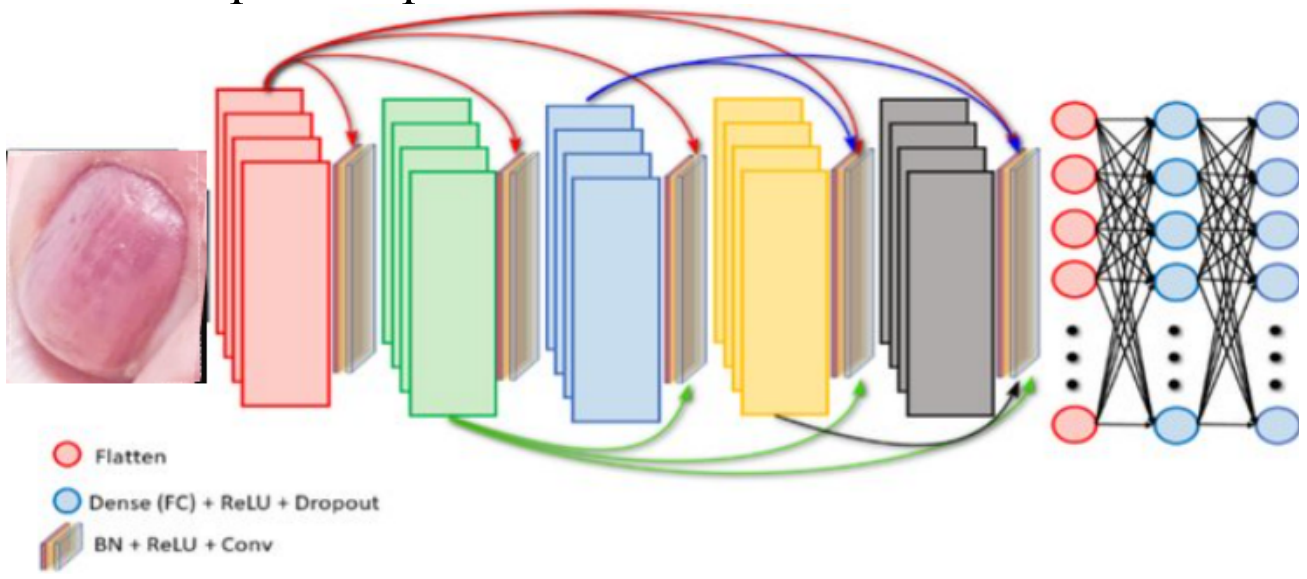

Fig. 4. Proposed DenseNet201 architecture

## C. EfficientNetV2

EfficientNetV2 is a strong convolutional neural network built to deliver high accuracy while training quickly. Fig. 5 shows, the uses two key blocks: Fused-MBConv for early layers and MBConv for deeper layers. Fused-MBConv simplifies operations to speed up learning, while MBConv includes feature expansion, depthwise convolution, squeeze-and-excitation (SE) to highlight important features, and channel reduction. The model uses Swish activation instead of ReLU to improve learning. It ends with global average pooling and a SoftMax layer to classify nail disease types accurately and efficiently.

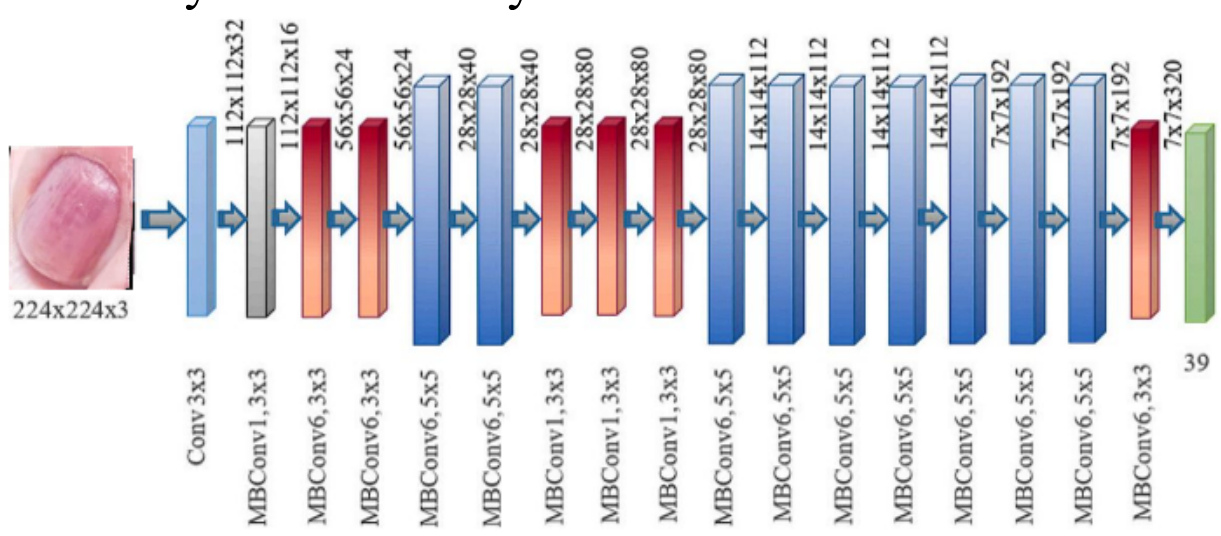

Fig. 5. Proposed EfficientNetV2 architecture

## D. ResNet50

ResNet50 is a deep neural network that has 50 layers and about 23 million parameters. It is also coordinated into five main parts called stages. Fig. 6 shows, each stage has several blocks, and each block has three convolution steps: first reduces size with 1×1 filters, second extracts feature with 3×3 filters, and third restores size with 1×1 filters. Batch normalization and ReLU activation are applied after each step. While the number of channels grows from 64 to 2048, the image size becomes smaller through stride-2 convolution. At the finishing stage, global average pooling turns features into one vector, which a fully connected layer with SoftMax uses to classify six nail disease types.

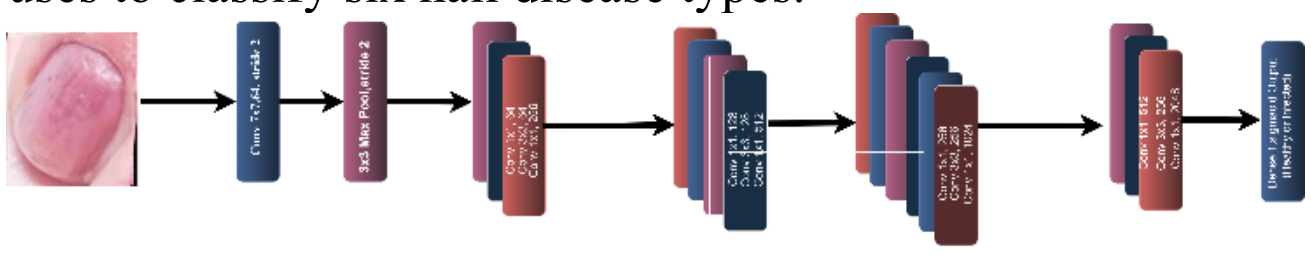

Fig. 6. Proposed ResNet50 architecture

# IV. EXPERIMENTAL ANALYSIS

## A. DenseNet201

Fig. 7 a noticeable performance improvement is observed in the early stages of training, where DenseNet201 attains nearly 99% accuracy on the training set and levels off at about 93% on the validation set. The model achieves high accuracy within the first 10 epochs and maintains consistent performance across the 70-epoch training cycle. While the training accuracy reaches near-perfect levels, the slight variance in validation accuracy (with a temporary dip around epoch 50) indicates minor fluctuations in generalization, which is expected in real-world datasets. Fig. 7 This graph shows how much error the model made during training. A lower loss means better performance. Both training and validation loss decreased fast and became very small after about 30 epochs.

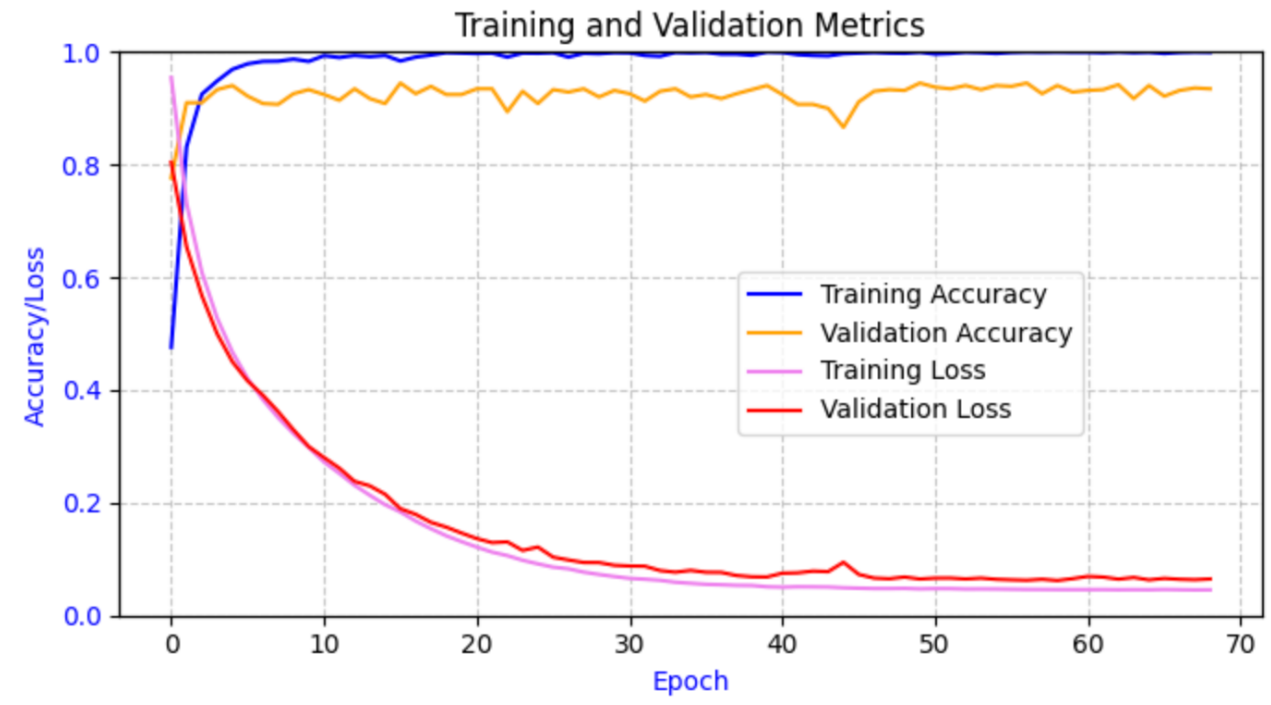

Fig. 7. Loss for training and validation

Fig. 8 below represents the confusion matrix of the DenseNet201 model. The model achieved high accuracy across most categories. Specifically, the model correctly classified 74 instances of Acral Melanoma, 45 of Healthy nails, 64 of Onychomycosis, 56 of Blue Fingernail, 74 of Clubbing, and 51 of Pitting. The confusion between certain classes, such as Pitting being misclassified as Onychomycosis (4 instances) and Clubbing (4 instances), highlights the visual similarity among these conditions. Similarly, 3 samples of Clubbing were misclassified as Onychomycosis, and a few Healthy samples were misclassified as Acral Melanoma (2 instances), suggesting the need for enhanced feature representation in borderline cases. Despite these minor misclassifications, the model demonstrates robust performance, particularly in correctly distinguishing between clinically distinct conditions.

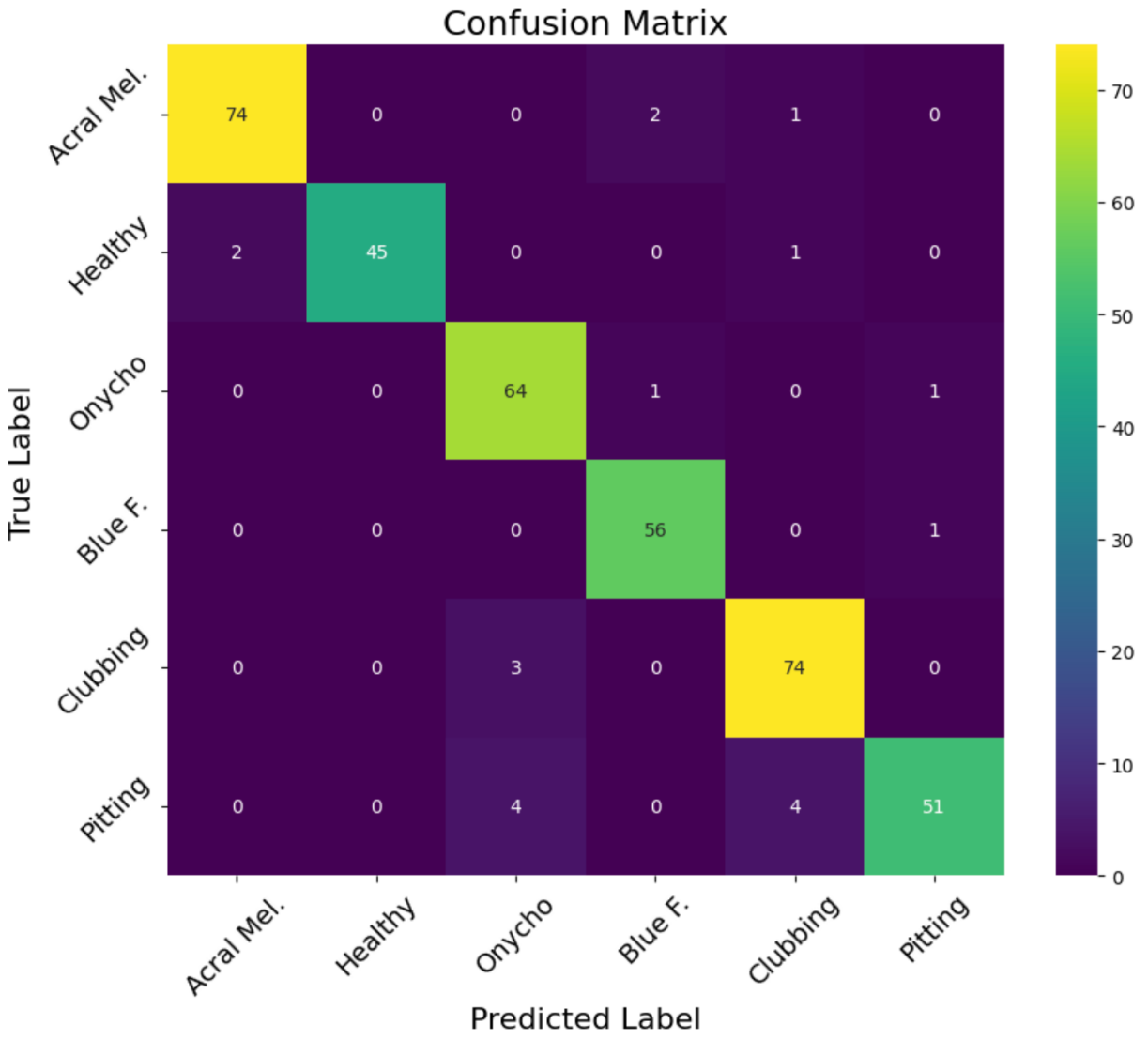

Fig. 8. Confusion Matrix of DenseNet201

The DenseNet201 model was trained with a batch size of 32, an initial learning rate of 0.0001, using the Adam optimizer and categorical_crossentropy loss function that shown in the table 1. We explored different combinations of learning rates (0.1, 0.01, 0.001, and 0.0001) and batch sizes (16, 32, and 64) to find the ideal combination. By training the models and monitoring their accuracy on an independent validation set, we chose the setup that achieved the highest overall performance. The highest validation accuracy reached was 93.47%, which helped ensure a strong and steady convergence.

TABLE I HYPERPARAMETER TUNING DENSENET201

| | | | |
|---|---|---|---|
| **Batch size** | 32 | **Loss function** | categorical_crossentropy |
| **Learning rate** | 0.0001 | **Number of epochs** | 69 |
| **Optimizer** | Adam | **Patience** | 10 |

Table 2 shows, DenseNet201 demonstrated even more balanced and robust performance across the same set of classes. It achieved perfect precision (1.00) for Healthy Nail and high precision for Pitting (0.96), Clubbing (0.93), and Blue Finger (0.95).

TABLE II CLASSIFICATION REPORT OF DENSENET201

| **Model Name** | **Classes** | **Precision** | **Recall** | **F1-Score** |
|---|---|---|---|---|
| DenseNet201 | Acral Lentiginous Melanoma | 0.97 | 0.96 | 0.96 |
| | Healthy Nail | 1.00 | 0.94 | 0.96 |
| | Onychogryphosis | 0.90 | 0.97 | 0.93 |
| | Blue Finger | 0.95 | 0.98 | 0.96 |
| | Clubbing | 0.93 | 0.96 | 0.94 |
| | Pitting | 0.96 | 0.86 | 0.91 |

### B. *InceptionV3*

The Fig. 9 illustrates the training and validation accuracy curves, with the blue line indicating training accuracy and the orange line showing validation accuracy. Initially, the model started with low accuracy (around 40%), but it improved quickly, reaching over 97% training accuracy in fewer than 10 epochs. Validation accuracy also improved, stabilizing between 90% to 94%, indicating strong performance on the test data. While the model achieves close to 100% accuracy on the training set, its validation accuracy plateaus at a somewhat lower level. This small gap between training and validation accuracy (about 3-5%) is acceptable and suggests that minor overfitting might be present, but is not severe. Fig. 9. in the loss curve plot, the blue line corresponds to training loss, while the orange line indicates validation loss. The model runs over 200 epochs. Because of the early stop, it reaches over 80 epochs. At the beginning (Epoch 0), both training and validation losses were high (above 12 and 11, respectively), indicating the model had not yet learned any meaningful features. Over the next 10 to 20 epochs, both losses dropped sharply, showing rapid learning. This drop indicates that the model quickly adapted to the patterns in the training data.

After about Epoch 30, the training loss plateaued near 0, and validation loss remained steady just slightly above zero, implying the model was not overfitting or underfitting. The consistently low values for both losses suggest the model showed good adaptability when tested on unfamiliar inputs.

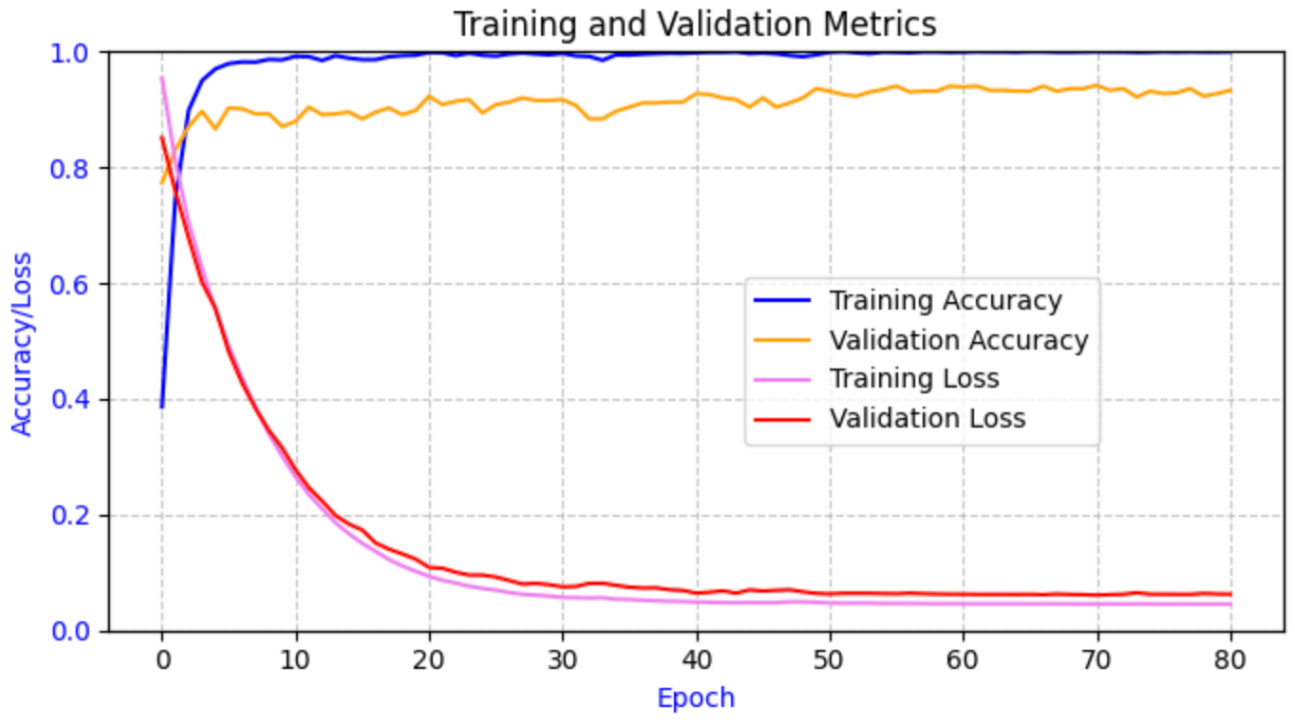

Fig. 9. Training and validation accuracy and loss curve

Fig. 10 below represents the confusion matrix of the InceptionV3. The InceptionV3 method demonstrated strong predictive capabilities, with most predictions aligning accurately with the true labels. Notably, the model achieved perfect classification for the “Pitting” class (59/59), and over 93% accuracy for other classes like “Onycholysis” (63 correct), “Blue Fingernail” (56 correct), and “Clubbing” (71 correct). Minimal misclassifications (1–4 per class) were observed, primarily between visually similar classes such as “Blue F.” and “Onycho.” Overall, the matrix indicates high generalization and class discrimination performance by the InceptionV3 architecture.

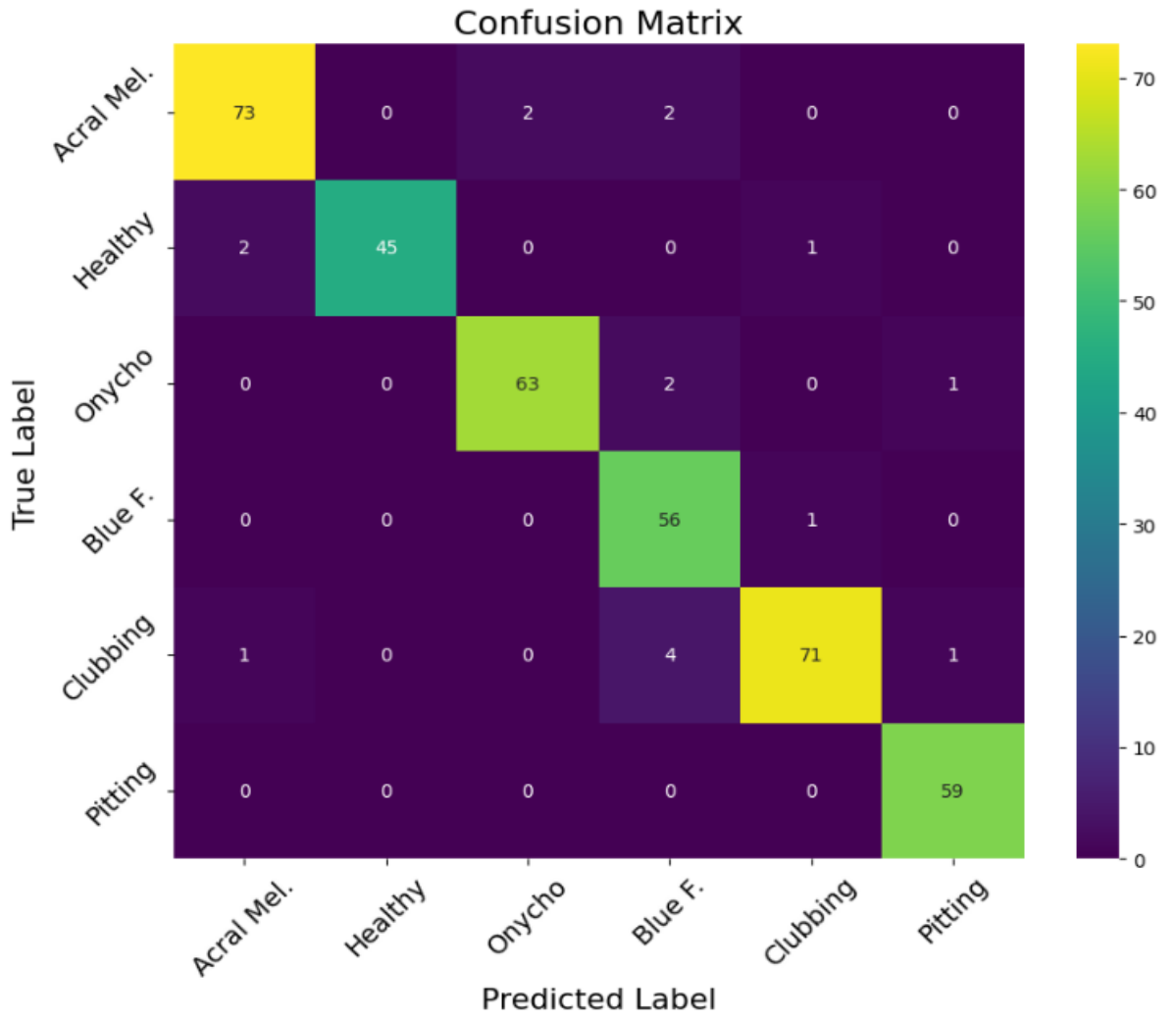

Fig. 10. Confusion Matrix of InceptionV3

In this model, we tested various learning rates (0.1 to 0.0001) and batch sizes (16, 32, 64) to optimize our model. The configuration that produced the highest accuracy on the validation set was selected. The proposed model was trained using a batch size of 32, a primary learning rate of 0.0001, and the Adam optimizer. The loss function applied was categorical_crossentropy, with a maximum of 200 epochs and an early stopping patience of 15. The best validation accuracy achieved during training was 93.32%. These settings contributed to stable convergence and effective model performance, which is shown in Table 3.

TABLE III HYPERPARAMETER TUNING OF INCEPTIONV3

| | | | |
|---|---|---|---|
| **Batch size** | 32 | **Loss function** | categorical_crossentropy |
| **Learning rate** | 0.0001 | **Number of epochs** | 81 |
| **Optimizer** | Adam | **Patience** | 10 |

Table 4 shows, the InceptionV3 model achieved strong classification results across all six nail disease categories. It achieved perfect precision (1.00) for Healthy Nail and high precision values for Pitting (0.97), Clubbing (0.97), and Onychogryphosis (0.97). However, it showed relatively lower precision in detecting Blue Finger (0.83), showing infrequent misclassifications in that class. Recall values remained consistently high, with perfect recall for Pitting (1.00) and strong results for the other classes, such as Acral Lentiginous Melanoma (0.95) and Onychogryphosis (0.95); however, a slightly lower recall was observed for Healthy Nail (0.94).

TABLE IV CLASSIFICATION REPORT OF INCEPTIONV3

| Model Name | Classes | Precision | Recall | F1-Score |
|---|---|---|---|---|
| InceptionV3 | Acral Lentiginous Melanoma | 0.96 | 0.95 | 0.95 |
| | Healthy Nail | 1.00 | 0.94 | 0.96 |
| | Onychogryphosis | 0.97 | 0.95 | 0.96 |
| | Blue Finger | 0.88 | 0.98 | 0.92 |
| | Clubbing | 0.97 | 0.92 | 0.93 |
| | Pitting | 0.97 | 1.00 | 0.98 |

Table 5, shows adversarial training results where different levels of adversarial noise, controlled by epsilon (ε), were applied to test how well the model could handle small input disturbances. Among all values, ε = 0.14 gave the best results, reaching the highest accuracy of 95.94% and lowest validation loss of 0.2031 in just 10 epochs. While ε = 0.18 and 0.2 also gave strong accuracy scores of 94.63%, their performance was marginally less stable. However, a higher epsilon, like 0.16, caused the model to perform poorly, showing a larger loss and stopping after only 2 epochs. These results show that adding a small to moderate amount of noise during training can improve the model's accuracy and help it become more reliable.

TABLE V PERFORMANCE OF INCEPTIONV3 FOR ADVERSARIAL EXAMPLES

| Epsilon Value (ε) | Validation Loss | Validation Accuracy | Optimal Epochs |
|---|---|---|---|
| 0 | 0.2302 | 0.9492 | 15 |
| 0.1 | 0.2322 | 0.9448 | 59 |
| 0.12 | 0.2273 | 0.9434 | 24 |
| 0.14 | 0.2031 | 0.9594 | 10 |
| 0.16 | 0.2792 | 0.9419 | 2 |
| 0.18 | 0.2202 | 0.9463 | 25 |
| 0.2 | 0.2311 | 0.9463 | 14 |

## V. Explainable AI

### A. *SHAP*

This Fig. 11 presents the damaged part of the cross-section of the nail. To improve transparency and trust in our nail disease classification model, we employed Explainable AI (XAI) methods like Grad-CAM, which generates heatmaps to pinpoint key regions in the nail images that impacted the model's predictions, helping us verify that the model focuses on relevant features like discoloration or texture. SHAP mainly highlights important features and assigns values to the features. This also explains how much the feature impacts the model's decision. It provides a deeper understanding by showing how each image feature contributes to the final prediction. Together, these methods make the model's decisions more understandable and reliable for medical use.

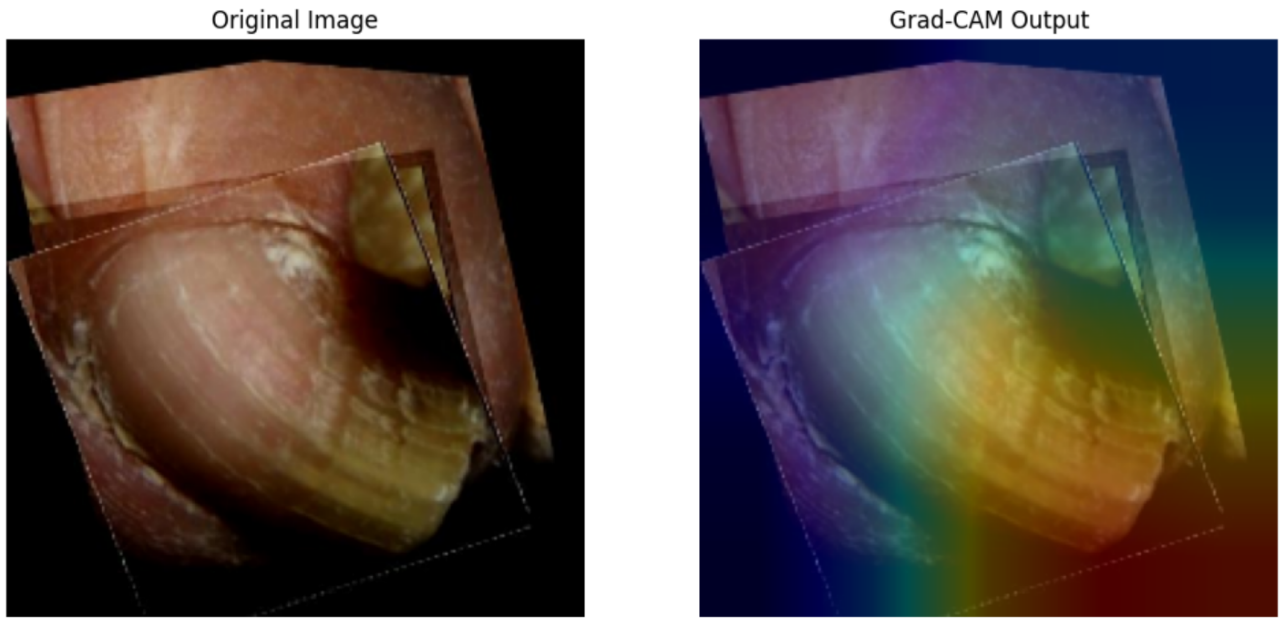

Fig. 11. Abnormality Identification using SHAP

## VI. Comparison

Among the models tested, InceptionV3 and DenseNet201 gave the best results for nail disease classification. As shown in Fig. 12, the models performed well across all six classes, but InceptionV3 slightly outperformed DenseNet201 overall. InceptionV3 achieved a higher average accuracy of 95.57%, while DenseNet201 reached 94.79%. In terms of class-wise performance, InceptionV3 showed stronger results for "Blue Finger" and "Pitting" with higher F1-scores, indicating better balance between precision and recall.

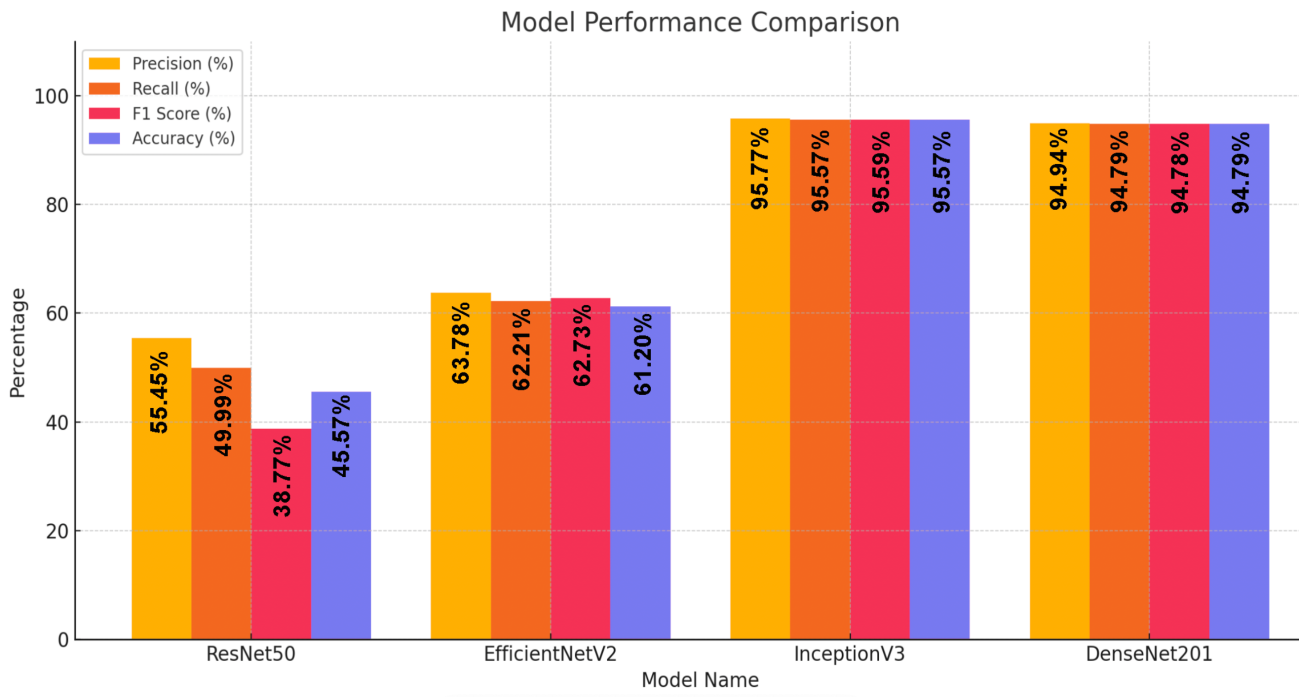

Fig. 12. Comparison between the four architectures

Table 6 presents a comparison between InceptionV3 and DenseNet201 architectures. Both models obtained an ideal training accuracy of 100%. On the test set, InceptionV3 had a small advantage over DenseNet201 with 95.57% vs 94.79%. However, DenseNet201 had a narrowly higher validation accuracy 93.47% compared to InceptionV3 93.32%, denoting powerful generalization for both models.

TABLE VI ACCURACY COMPARISON BETWEEN THE TWO ARCHITECTURES

| Algorithm | InceptionV3 | DenseNet201 |
|---|---|---|
| Training accuracy | 100.00% | 100.00% |
| Test accuracy | 95.57% | 94.79% |

Table 7 presents a comparison of different deep learning models used by previous researchers and the models proposed in this study. Ardianto et al. [11] used a basic CNN model and achieved 83% accuracy. Kaur et al. [12] applied DenseNet121 and got 81.3% accuracy. Palloge et al. [13] tested multiple AE-based models. Their AE-CNN gave the best result at 91.8%, while AE-VGG16 achieved 83.5%, and AE-ResNet50 had the lowest accuracy at 48.3%. In contrast, our proposed models, InceptionV3 and DenseNet201, outperformed all the previous methods. InceptionV3 achieved the highest accuracy of 95.57%, and DenseNet201 also performed strongly with 94.79% accuracy. These results highlight the effectiveness of our proposed deep learning approaches for nail disease classification

TABLE VII COMPARISON WITH PREVIOUS WORKS

| Author | Method | Accuracy (%) |
|---|---|---|
| Ardianto et al. [11] | CNN | 83 |
| Kaur et al. [12] | DenseNet121 | 81.3 |
| Palloge et al. [13] | AE-CNN | 91.8 |
| | AE-VGG16 | 83.5 |
| | AE-ResNet50 | 48.3 |
| Proposed Model | DenseNet201 | 94.79 |
| | InceptionV3 | 95.57 |

## VII. Conclusion

Nails, one of the important parts of our body, may be infected by different kinds of diseases. These diseases may cause fatal consequences to our bodies. The study focuses on applying deep learning approaches to detect and classify nail disorders through image analysis. To classify six types of nail conditions: acral lentiginous melanoma, healthy nail, onychogryphosis, blue finger, clubbing, and pitting, four CNN models- DenseNet201, EfficientNetV2, InceptionV3, and ResNet50 are used in this study. Among the models, InceptionV3 has the highest accuracy of 95.57%. This model has demonstrated superior feature extraction and generalization capabilities. In terms of DenseNet201, it achieved 94.79% accuracy. Sometimes, the early signs of nail disease are overlooked or misdiagnosed. The outcomes of the study highlight the capability of AI-based systems in early and accurate diagnosis of nail diseases. By using color-based features and high-performing CNN models, this study may help to develop non-invasive, cost-effective diagnostic tools. There are some limitations of the study that may be mitigated in the future. To train the models, we used a dataset with 3,835 images. The number of images is relatively small to get the best output. It may affect the generalizability of the models. Besides, the image quality was not very good. This can be a reason to lessen the accuracy of the models. In the future, this study can be extended using a comparatively large number of images, which may improve the generalization capability of the model. However, this model can be useful to dermatologists and general healthcare providers in early disease detection.

## References

[1] Regin, R., Reddy, G. G., Kumar, C. S., & CVN, J. (2022). Nail Disease Detection and Classification Using Deep Learning. Central Asian Journal of Medical and Natural Science, 3(3), 574-594.

[2] Bang, S., Magarde, A., Chandak, D., Agrawal, K., & Morankar, G. (2023). Nail Disease Detection Using AI-Based Algorithm. In N. Chaki, N. D. Roy, P. Debnath, & K. Saeed (Eds.), Proceedings of International Conference on Data Analytics and Insights, ICDAI 2023 (pp. 355-364). Springer, Singapore.

[3] Thakur, A., & Kaur, A. (2023). Nail Disease Classification Using Convolutional Neural Network and Transfer Learning. 2023 International Conference on Computing, Communication, and Intelligent Systems (ICCCIS), 1014-1019. IEEE.

[4] Kemenes, S., Chang, L., Schlereth, M., Noversa de Sousa, R., Minopoulou, I., Fenzl, P., Corte, G., Yalcin Mutlu, M., Höner, M. W., Sagonas, I., Coppers, B., Liphardt, A.-M., Simon, D., Kleyer, A., Folle, L., Sticherling, M., Schett, G., Maier, A., & Fagni, F. (2025). Advancement and independent validation of a deep learning-based tool for automated scoring of nail psoriasis severity using the modified nail psoriasis severity index. Frontiers in Medicine, 12, 1574413.

[5] Horikawa, H., Tanese, K., Nonaka, N., Seita, J., Amagai, M., & Saito, M. (2024). Reliable and easy-to-use calculating tool for the Nail Psoriasis Severity Index using deep learning. NPJ Systems Biology and Applications, 10(130).

[6] Folle, L., Fenzl, P., Fagni, F., Thies, M., Christlein, V., Meder, C., Simon, D., Minopoulou, I., Sticherling, M., Schett, G., Maier, A., & Kleyer, A. (2023). DeepNAPSI multi-reader nail psoriasis prediction using deep learning. Scientific Reports, 13, 5329.

[7] Jansen, P., Creosteanu, A., Matyas, V., Dilling, A., Pina, A., Saggini, A., Schimming, T., Landsberg, J., Burgdorf, B., Giaquinta, S., et al. (2022). Deep learning assisted diagnosis of onychomycosis on whole-slide images. Journal of Fungi, 8(9), 912

[8] Indi, T. S., & Gunge, Y. A. (2016). Early stage disease diagnosis system using human nail image processing. International Journal of Information Technology and Computer Science, 7(7), 30–35

[9] Hamim, M. A., Rupak, A. U. H., Hasan, M. S., & Ray, K. (2023). Multiple nail-disease classification based on machine vision using transfer learning approach. In 2023 International Conference on Computing, Communication and Networking Technologies (ICCCNT) (pp. 1–6). IEEE

[10] Gurav, N. (2024). Nail Disease Detection Dataset. Kaggle. https://www.kaggle.com/datasets/nikhilgurav21/nail-disease-detection-dataset

[11] Ardianto, R., Yusuf, D., Sumantri, R. B. B., Febrina, D., Al-Hakim, R. R. R., & Ariyanto, A. S. S. (2025). Bioinformatics-driven deep learning for nail disease diagnosis: A novel approach to improve healthcare outcomes. BIO Web of Conferences, 152, 01024.

[12] Kaur, A., & Gurrapu, N. (2025). Smart diagnosis: Transforming nail disease identification with deep learning. In 2025 International Conference on Next Generation Communication & Information Processing (INCIP) (pp. 807–812). IEEE.

[13] Palloge, A. H., Tahir, Z., & Syafaruddin. (2025). Comparative analysis of multiclass classification using AE-CNN method combination on nail diseases. In 2025 International Conference on Advancement in Data Science, E-learning and Information System (ICADEIS) (pp. 1–7). IEEE.